\documentclass[runningheads]{llncs}
\usepackage[T1]{fontenc}
\usepackage{graphicx}
\usepackage{hyperref}
\usepackage{color}

\usepackage{multirow}
\usepackage{booktabs}
\usepackage{adjustbox}
\usepackage{tabularx}
\usepackage{longtable, tabu}
\usepackage{subcaption}
\usepackage{pifont}
\usepackage{tablefootnote}
\usepackage{threeparttable}
\usepackage{CJKutf8}

\usepackage{xcolor}

\begin{document}

\title{SmartPhone: Exploring Keyword Mnemonic with Auto-generated Verbal and Visual Cues}
\titlerunning{SmartPhone}
\author{Jaewook Lee \and Andrew Lan}
\authorrunning{J. Lee et al.}
\institute{University of Massachusetts Amherst \\
Contact Email: \email{jaewooklee@cs.umass.edu}\\
}
\maketitle              % typeset the header of the contribution
\begin{abstract}
In second language vocabulary learning, existing works have primarily focused on either the learning interface or scheduling personalized retrieval practices to maximize memory retention. However, the learning content, i.e., the information presented on flashcards, has mostly remained constant. Keyword mnemonic is a notable learning strategy that relates new vocabulary to existing knowledge by building an acoustic and imagery link using a keyword that sounds alike. Beyond that, producing verbal and visual cues associated with the keyword to facilitate building these links requires a manual process and is not scalable. In this paper, we explore an opportunity to use large language models to automatically generate verbal and visual cues for keyword mnemonics. Our approach, an end-to-end pipeline for auto-generating verbal and visual cues, can automatically generate highly memorable cues. We investigate the effectiveness of our approach via a human participant experiment by comparing it with manually generated cues.

\end{abstract}
\section{Introduction}
Learning vocabulary is key to learning second (mostly foreign) languages, but also a difficult task. One of the most well-known and effective methods is flashcards, i.e., writing the L2 (a second language word) word on the front and writing down the corresponding L1 word (a first or native language word) on the back, with content such as mnemonic or context. Moreover, one may manage flashcards by putting the cards in boxes to follow the Leitner system~\cite{leitner1974so} to recall the word regularly following the forgetting curve~\cite{ebbinghaus2013memory}. However, both writing down every word and managing a bunch of cards require significant effort and can take a lot of effort from learners. 

Technology advances have enabled vocabulary learning to shift from manually writing down the words to using software systems such as Anki~\cite{anki} and Quizlet~\cite{quizlet}, which make language learning more efficient and engaging. Some systems use ideas behind intelligent tutoring systems to model the learner's knowledge state to intervene in the retrieval practice~\cite{reddy2016unbounded,ye2022stochastic,zylich2021linguistic}. Many studies have shown that managing retrieval practice and designing personalized schedules using cognitive models can significantly improve learning efficiency~\cite{carrier1992influence,larsen2009repeated}. Many systems also use gamified interfaces and enable learners to share decks with others, making the learning process more interactive and socially relevant~\cite{duolingo,anki,quizlet}. However, despite these advances, the learning \emph{content}, i.e., what is written on the flashcard, has mostly stayed the same throughout the years. 

Regarding the content for second language learning, keyword mnemonic~\cite{atkinson1975application} is a notable memory encoding strategy that uses interactive visual imagery with a keyword that sounds like part of a foreign word. Forming the keyword-based interactive image takes a two-step approach: creating first an acoustic and then an imagery link. Imagine a native English speaker is learning the Spanish word \textit{pato}, which means \textit{duck}. The keyword that sounds like the word is \textit{pot}. Using the keyword, the learner first creates an acoustic link between the keyword and the Spanish word. Then, the learner builds an imagery link that connects the sound and its meaning by using a verbal cue, such as ``A duck wearing a pot on its head.'' By relating new information to existing knowledge, learners have an easier time memorizing the word and can retain it in memory for a longer time.

Previous studies on keyword mnemonics have shown their effectiveness compared with different learning strategies. Comparing keyword mnemonic with rote rehearsal and combining both strategies showed that the keyword group outperformed the other two groups~\cite{brahler2008learning}. Comparing the keyword mnemonic group with verbal and visual cues with mixed methods of contextual clues, word structure analysis, and opposite word pairs showed that the keyword group performed better in both short-term and long-term retention~\cite{siriganjanavong2013mnemonic}. However, since the cues given in these studies are manually generated by experts, it is difficult to employ this approach at a large scale in the systems mentioned above.

In 2014, Savva et al.\ introduced an automatic keyword generation approach based on a cross-lingual system, TransPhoner~\cite{savva2014transphoner}. It evaluates candidate keywords in the second language using the following measures for a given input word: imageability, phonetic similarity, orthographic similarity, and semantic similarity. The authors experimented on the effectiveness of TransPhoner using an evaluation set of 36 German words~\cite{ellis1993psycholinguistic} with three other conditions: no keywords, randomly sampled keywords, and manually generated keywords. The result shows that the TransPhoner-generated condition achieved the highest score and the manually-generated keyword condition had no significant difference from randomly generated keywords.

Despite TransPhoner's success in automatically generating keywords as cues, other forms of richer verbal or visual cues that could further help learners build an imagery link cannot be automatically generated. The learner (or teacher) still needs to manually develop them to connect the keyword and the L1 word, which requires a lot of effort on their part. Moreover, it takes an expert to come up with an image as the visual cue that corresponds to the verbal cue. Using image APIs such as Google Image API, one can juxtapose images of a keyword and an L1 word, but doing is not as effective as showing both words together in a single image. To make keyword mnemonic scalable, we need an end-to-end solution that takes words as input and generates keyword, verbal and visual cues. 

\textbf{Contributions.} In this paper, we detail a pipeline for automatically generating verbal and visual cues in one shot via text generator and text-to-image generator. Our contributions are as follows:
\begin{itemize}
  \item We propose a large language model (LLM)-based pipeline that automatically generates highly memorable verbal and visual cues for an L1 word in language learning. We believe that our automated approach will significantly reduce content development costs by enhancing time efficiency and reducing manual generation effort. To the best of our knowledge, we are the first to apply LLMs in the context of keyword mnemonic.
  \item We implement a web application for human participant studies and use it to compare our approach with existing ones. We analyze the effectiveness of four approaches: automatically generated keyword only, automatically generated keyword with a verbal cue, automatically generated keyword with both verbal and visual cues, and manually generated keyword and verbal cues. We also outline avenues for future work that could stem from our approach. 
\end{itemize}

\section{Methodology}
In this section, we detail our pipeline for automatically generating cues. Our work is driven by the following two research questions:

\begin{itemize}
  \item Can we automatically generate human-level verbal cues for the keyword?
  \item Can we generate a visual cue that may facilitate building an imagery link that is described in a verbal cue?
\end{itemize}

We narrow the scope of automatically generating verbal and visual cues to the experiments conducted in previous studies~\cite{ellis1993psycholinguistic,savva2014transphoner} in this preliminary effort. We use the evaluation set of 36 German words and keywords from previous studies for both manually and automatically generated cues as baselines. Since verbal cues only exist for manually generated keywords, our task boils down to automatically generating verbal cues using TransPhoner-generated keywords and generating visual cues using verbal cues.

\subsection{Pipeline for Auto-generating Verbal and Visual Cues}
\begin{figure}
\centering
\includegraphics[width=.8\textwidth]{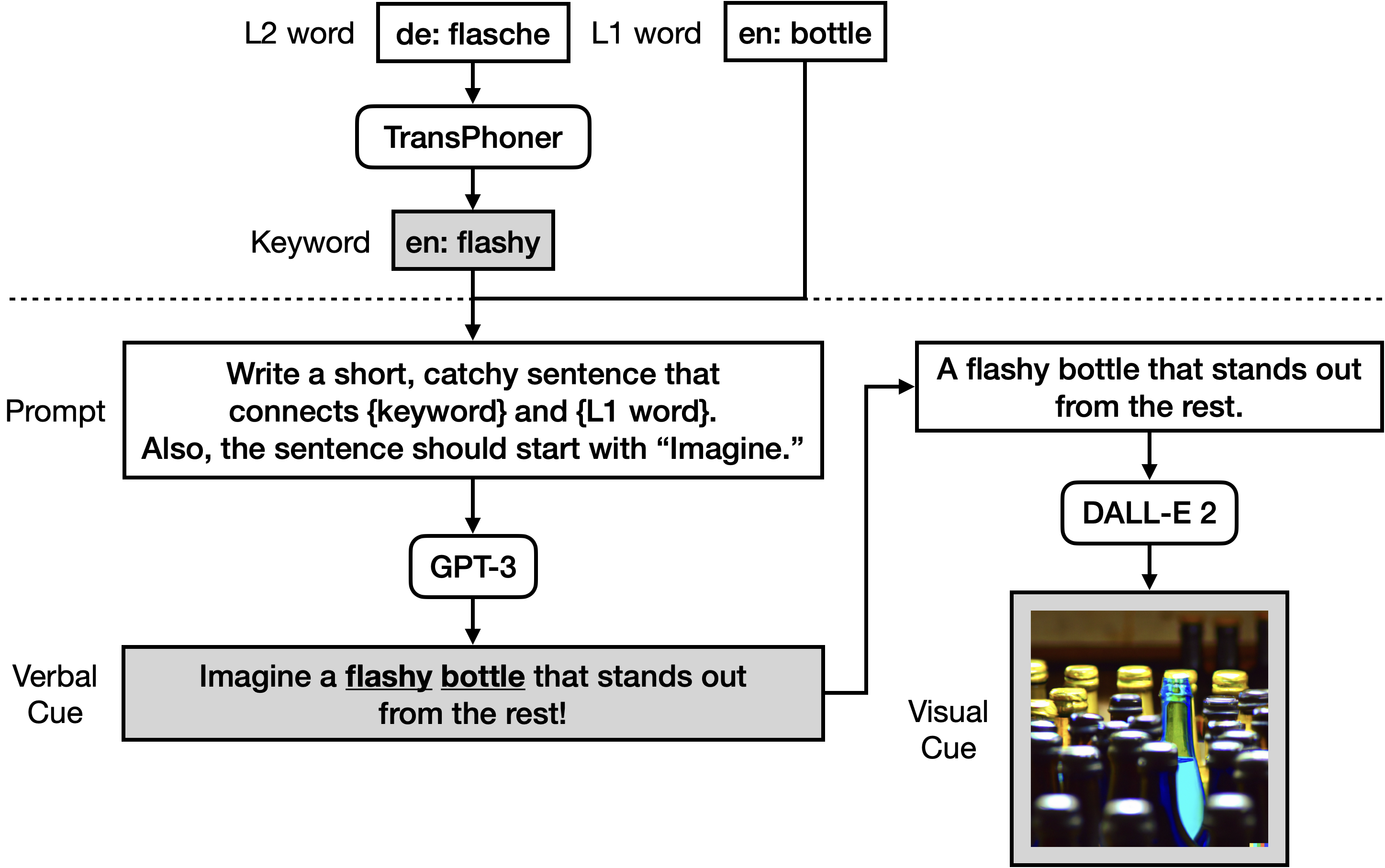}
\caption{Our end-to-end pipeline for automatically generating verbal and visual cues for an L2 word.} \label{fig:pipeline}
\vspace{-.5cm}
\end{figure}

We propose a pipeline consisting of two LLMs that generate verbal and visual cues in two steps: First, we use a text generator to automatically generate a sentence containing the TransPhoner keyword as the verbal cue. Second, we use a text-to-image generator to generate an image as the visual cue. 
LLMs, pre-trained with massive datasets, have shown human-level performance on the tasks we described above through prompts. This is because LLMs are good for controllable text generation~\cite{prabhumoye2020exploring} and following instructions~\cite{ouyang2022training}. With proper prompts, models show their ability to solve the tasks with zero-shot or few-shot setups. We use zero-shot setup LLMs for both generating verbal and visual cues. 
We detail the pipeline through an example in Fig.~\ref{fig:pipeline} where we need to generate cues for the German word \textit{flasche}, which means a \textit{bottle}. 
The keyword generated by TransPhoner is \textit{flashy}; Using the keyword and the meaning of the word, we create the prompt: ``Write a short, catchy sentence that connects \textit{flashy} and \textit{bottle}.'' Additionally, we add a constraint on verbal cues to start with ``Imagine'' for two reasons. First, verbal cues in the previous study~\cite{ellis1993psycholinguistic} are in that format. Since we are trying to answer whether we could achieve human-level verbal cues, we match the format. Second, we follow grammatical characteristics that come after the word imagine. After the word ``Imagine'', usually a noun or gerund comes out; we found that the generated verbal cue contains fewer ambiguous pronouns, which makes the cue more descriptive. This feature is key to linking the text generator and text-to-image generator within the same pipeline. 
% \sloppy
Using the prompt, our text generator, GPT-3~\cite{brown2020language} (text-davinci-003, temp=0.5), generates the verbal cue. Then, we reuse the verbal cue as the prompt for our text-to-image generator, DALL-E 2~\cite{dalle}, by removing the word ``Imagine''. One can freely choose any LLMs to automatically generate these verbal and visual cues. We present the gray region in Fig.~\ref{fig:pipeline} to the participant as learning content. 

\section{Experimental Evaluation}

In this section, we detail our experiments on presenting different content to different participants to explore whether automatically generated verbal and visual cues are effective in vocabulary learning. 

\subsection{Experimental Design}

In the experiment, participants learn 36 German words and are tested on recalling both the German word (generation) and its English meaning (recognition). The words are split into three sets, which means that each participant goes through the learning, recognition, and generation cycle three times. Words in each set are randomly shuffled for each participant. At the end of the experiment, we also ask participants to rate the helpfulness of the cues. 

\subsubsection{Learning and Testing}

We provide each participant with both instructions on the study and the content that helps them learn the word; see Section~\ref{sssec:cond} for details. Each word has a 30-second limit for the participant to memorize, and the participant can choose to move on to the next word after 15 seconds. After 30 seconds, we automatically move on to the next word. German words are pronounced twice, after 2 seconds and 7 seconds, respectively, after being displayed. We show a timer to participants to make them aware of the time remaining for each word. 
%
%\subsubsection{Testing}
%
Participants have 15 seconds for both recognition and generation during testing. To avoid confusion between the two tests, we provide instructions such as ``What is this in English?'' and ``What is this in German?''. For generation, we also ask participants to use \textit{a, o, u, s} instead of Umlaut \textit{{\"a}, {\"o}, {\"u}, {\ss}}. We show a timer to participants as well. Words in both tasks are randomized in order. 

\subsubsection{Participants}

We recruit participants from Amazon Mechanical Turk~\cite{amt}. We require participants to be native English speakers with no German language experience. Considering the experiment takes about 40 minutes, we paid each participant \$7.25 and added a bonus of \$2.75 for those who got a score of over 70\% on the final test. The bonus encourages participants to do their best. However, we acknowledge that some participants may cheat on tests to achieve a high score by using an external dictionary, which we have no control of.

\subsubsection{Web Interface}

We implement a React web application as our participant interface, which is designed based on the previous study~\cite{savva2014transphoner}. We place an IRB-approved consent form on the front page and only participants who agree can participate in the experiment; the form explains in detail how the experiment is structured. We also show an example with a German word not in our evaluation set to clarify the procedure to participants. We collect metadata on time spent both during learning and testing, along with the responses to further investigate participant behavior.

\subsection{Experimental Conditions} \label{sssec:cond}

We first divide participants into two groups based on how the keyword was generated: automatically (auto-cue) and manually (manual-cue). Among many combinations of verbal and visual cues that can be presented to the participants, we choose conditions that enable both intra- and inter-group comparisons. We recruit a total of 80 participants for our study, with 20 in each condition.

\begin{figure}[t]
\centering
\includegraphics[width=.8\textwidth]{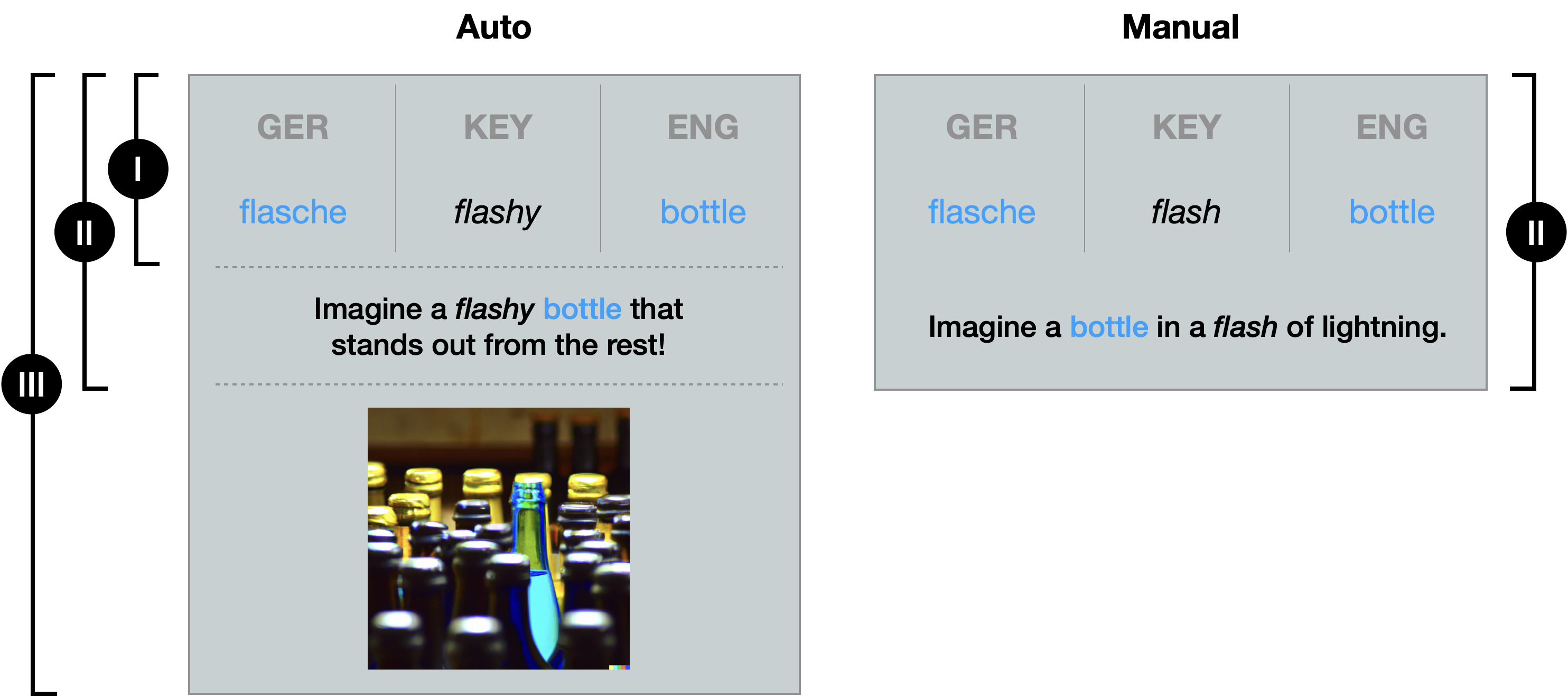}
\caption{A snapshot of our web interface shown to experiment participants.} \label{fig:interface}
% \vspace{-.7cm}
\end{figure}

As shown in Fig.~\ref{fig:interface}, we show the example of our web interface on how the content is displayed in different conditions. For intra-group comparisons, we further divide the auto-cue group into three conditions: Condition I is only provided with the TransPhoner-generated keyword, Condition II is provided with the keyword and the verbal cue generated by our pipeline, and Condition III is provided with the keyword and both the verbal and visual cues generated by our pipeline. For the inter-group comparisons, we provide both the auto-cue group and manual-cue group with information in Condition II. We note that the previous study~\cite{savva2014transphoner} compared the groups with Condition I by not including verbal cues that were originally presented with the manually generated keywords~\cite{ellis1993psycholinguistic}. The manually generated verbal cue and keyword should be considered as a whole since the keyword might have been chosen to provide a verbal cue with the best imageability among many keyword candidates. 

We refer to these four conditions as Auto-I, Auto-II, Auto-III, and Manual-II. The instructions for each condition are shown in Table~\ref{tab:cond_instr}. We use the same instructions for Condition I from Savva et al. Our instructions for Condition II tell participants to create an imagery of a scene specified in a verbal cue. Our instructions for Condition III tell participants to remember the image, which is based on the verbal cue that describes a specific scene. 

% X[1,c]X[4,l]X[4,l]
\renewcommand\tabularxcolumn[1]{m{#1}}
\newcolumntype{b}{>{\hsize=2.3\hsize}X}
\newcolumntype{s}{>{\hsize=.45\hsize}>{\centering\arraybackslash}X}

\begin{table}[]
\caption{Cues and instructions we used for different experimental conditions.}
\begin{tabularx}{\textwidth}{ s s s s b }
\multirow{2}{*}{Cond.} & \multicolumn{3}{c}{Cue} &  \multicolumn{1}{c}{\multirow{2}{*}{Instruction}}  \\ \cline{2-4} & Keyword & Verbal & Visual \\ \hline
I & yes & no & no & Imagine a visual scene connecting the given keyword with the English meaning, and the sound of the German word. \\ \hline
II & yes & yes & no & Imagine a specific scene described in the verbal cue that connects the given keyword with the English meaning, and the sound of the German word. \\ \hline
III & yes & yes & yes & Remember the image by following the verbal cue that connects the given keyword with the English meaning, and the sound of the German word.  \\ \hline \\
\end{tabularx}
\label{tab:cond_instr}
\vspace{-.5cm}
\end{table}

\subsection{Evaluation Metrics}

We use different metrics to score recognition and generation. For recognition, we use cosine similarity between the word embeddings~\cite{bojanowski2017enriching} between the answer and the response. We also consider responses that miss ``to'' for ``to''-infinitives to be correct. Unlike recognition, as a novice German learner, generation is bounded to the orthographic feature of vocabulary. Therefore, we use a standardized (subtracting 1 and normalizing to 1) Levenshtein distance to score generation, following previous studies~\cite{savva2014transphoner}. 
We also ask participants to evaluate the helpfulness of the cues using a 5-point Likert scale, which is provided along with the entire 36 words and the cues. 

\subsection{Results and Discussion}

After we exclude participants who did not understand the experiment properly, such as those who wrote down the keyword when recalling the English meaning, we have a total of 72 participants: Auto-I (20) with an average age of 25.4 years (SD = 2.3), Auto-II (17) with an average age of 24.2 years (SD = 1.7), Auto-III (18) with an average age of 24.8 years (SD = 1.6), and Manual-II (17) with an average age of 25.3 years (SD = 1.1). 

\begin{figure}[t]
\includegraphics[width=\textwidth]{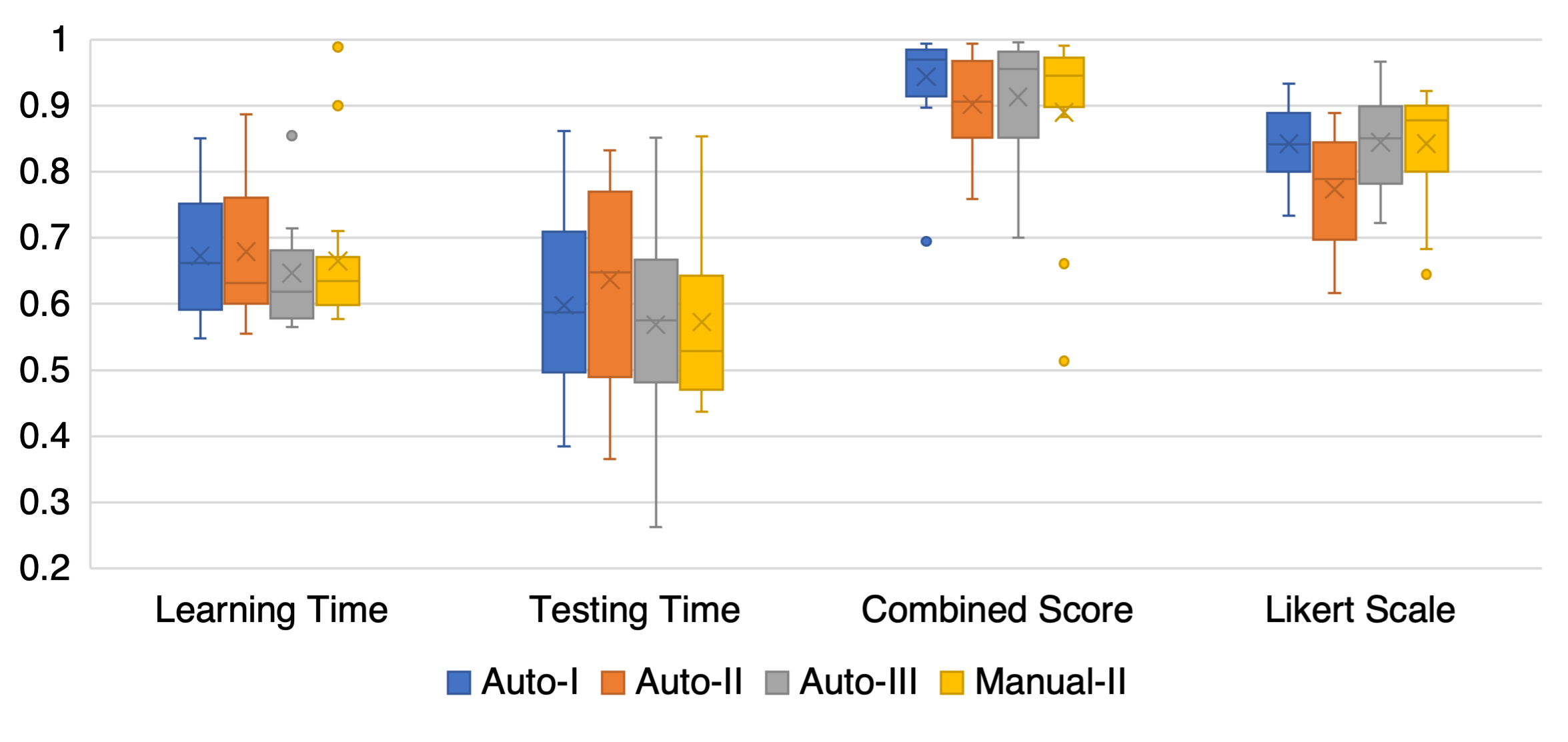}
% \vspace{-.7cm}
\caption{Box plots of per-participant data for each experimental condition.} \label{fig:box_plot}
\vspace{-.5cm}
\end{figure}

Fig.~\ref{fig:box_plot} shows per-participant experimental data in box plots averaged among 36 German words. Learning time is time spent memorizing a word, while testing time is the average time on recognition and generation of the word. Similarly, the combined score is an average of recognition and generation scores. Learning time, testing time, and Likert scale are normalized by their maximum value.

The median of time spent on learning was 19.8, 18.9, 18.6, and 19.2 seconds, respectively, for the four conditions out of the 30 seconds time limit, which may suggest that cognitive load across different conditions is similar. The median of time spent on testing, i.e., the average time spent on recognition and generation, was 8.85, 9.75, 8.7, and 7.95 seconds out of the 15 seconds time limit. The median of the 5-point Likert scale was 4.2, 3.95, 4.25, and 4.4. 

Now, we analyze the combined score based on the per-word combined score, as shown in Fig.~\ref{fig:combined_score}. We perform a one-tailed Welch's t-test assuming unequal variances on the hypotheses of one condition being better than another. We set our level of significance to 5\%. We detail each hypothesis below. Case A, B, and C in Fig.~\ref{fig:combined_score} are words we present with content generated through our pipeline for qualitative analysis.  

\subsubsection{Auto-I vs.\ Auto-II: Does a verbal cue help learning?} 
We hypothesize that Auto-II, with additional verbal cues, will result in better recognition and generation scores than Auto-I, which uses only keywords. We define our null hypothesis ($H_{0}$) and alternate hypothesis ($H_{a}$) as follows:

\begin{itemize}
  \item $H_{0}$: $\mu_{Auto-II} \leq \mu_{Auto-I}$
  \item $H_{a}$: $\mu_{Auto-II} > \mu_{Auto-I}$
\end{itemize}

A right-tailed test shows there is no significant effect of verbal cues, $t(33) = -1.79, p = 0.96$; we cannot reject $H_{0}$. On the contrary, a left-tailed test shows statistical significance in favor of the keyword-only condition, $t(33) = 1.79, p = 0.04$. This result can be explained by several factors: The participants might have done rote rehearsals instead of building links as instructed in Table~\ref{tab:cond_instr}. Moreover, participants may come up with their own verbal cues that are more memorable than automatically generated ones. Personalized by default, participants' own verbal cues may be a better fit for each individual's own experience.

\subsubsection{Auto-II vs.\ Manual-II: Are automated verbal cues effective?} 
We hypothesize Manual-II to be an upper bound of Auto-II since the former cues are generated by experts in psycholinguistics. Therefore, we define our null hypothesis and alternate hypothesis as follows:

\begin{itemize}
  \item $H_{0}$: $\mu_{Manual-II} \leq \mu_{Auto-II}$
  \item $H_{a}$: $\mu_{Manual-II} > \mu_{Auto-II}$
\end{itemize}

\begin{figure}[t]
\includegraphics[width=\textwidth] {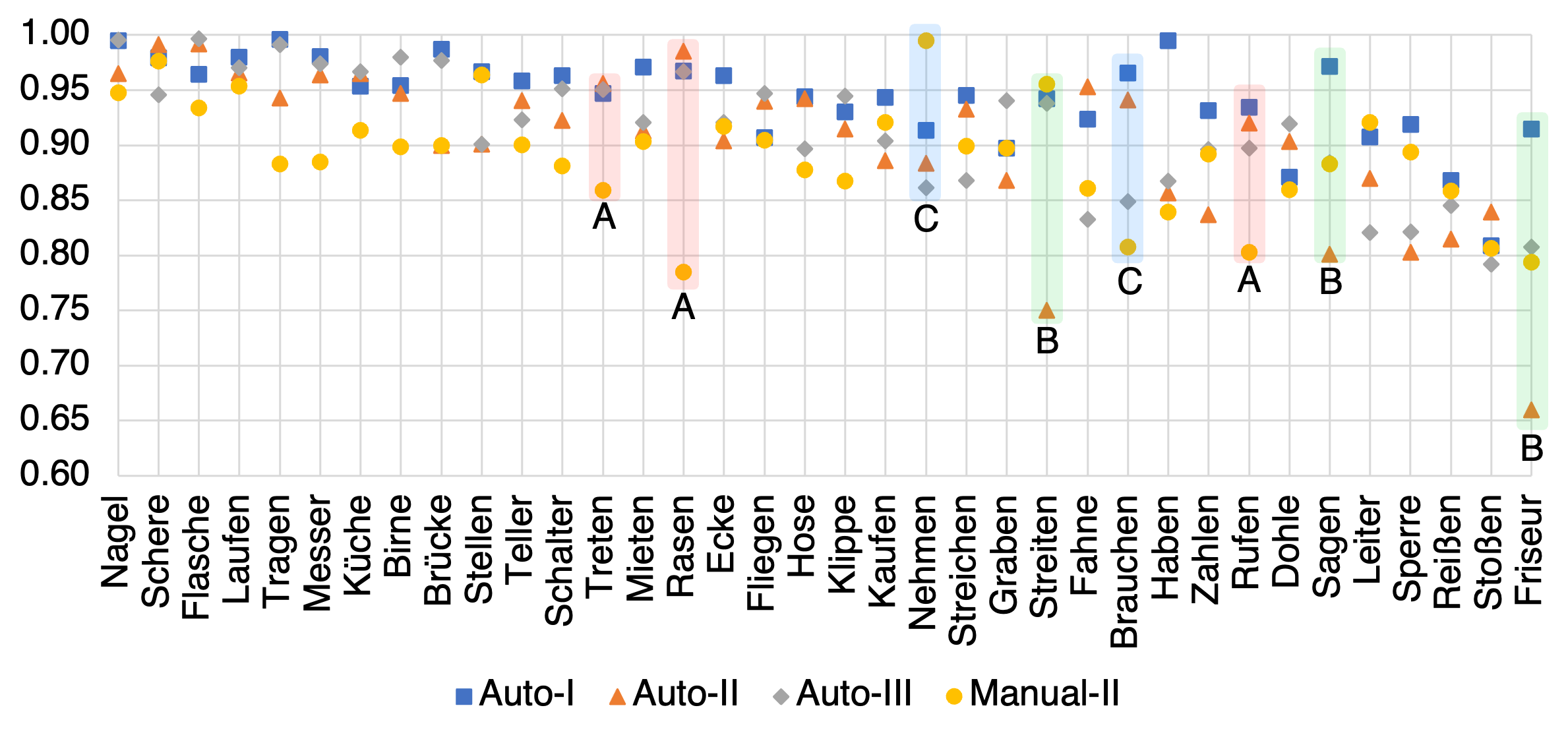}
% \vspace{-.5cm}
\caption{Per-word combined score for all four experimental conditions, with three cases highlighting some words that work especially well with certain cues.} \label{fig:combined_score}
\vspace{-.5cm}
\end{figure}

A right-tailed test shows that there is no significant difference between the two conditions, $t(24) = -0.32, p = 0.62$; we cannot reject $H_{0}$. In Fig.~\ref{fig:combined_score}, we show three words where participants perform better in the Auto-II condition than Manual-II (case A) and otherwise (case B), respectively. Case A in Table~\ref{tab:verbal_cue} shows that auto-generated cues are more memorable than manual cues even with a grammatical error (risen should be raised) or not realistic (Reuben sandwich calling your name). Case B, on the other hand, contains keywords where auto-generated cues are not frequently used (Triton, frizzy) or making it hard to imagine (a wagon with stories). This result implies that although we can automatically generate high-quality verbal cues, choosing appropriate keywords remains crucial. Therefore, we need to add keyword generation to the pipeline and evaluate the quality of both generated keywords and the verbal cue.

\begin{table*}[t]
\caption{Examples of automatically and manually generated verbal cues. A keyword is represented in \textit{italic}, while a meaning is in \textbf{bold}.}
\begin{tabularx}{\textwidth}{ccXX}
Case               & Word     & Auto                                                                                                                            & Manual                                                                                                        \\\hline
\multirow{3}{*}{A} & Treten   & Imagine \textbf{stepping} into \textit{treason}, a treacherous path that can never be undone. & Imagine you \textbf{step} on a stair \textit{tread}.                        \\
                   & Rasen    & Imagine a \textit{risen} \textbf{lawn} that is lush and green!                                & Imagine your \textbf{lawn} covered in \textit{raisins}.                     \\
                   & Rufen    & Imagine \textit{Reuben} \textbf{calling} out your name!                                       & Imagine you \textbf{call} a friend to put a new \textit{roof} on a cottage. \\\hline
\multirow{3}{*}{B} & Streiten & Imagine \textit{Triton} and his trident \textbf{quarreling} with the waves.                   & Imagine you \textbf{quarrel} about the Menai \textit{straits}.              \\
                   & Sagen    & Imagine a \textit{wagon} full of stories just waiting to be \textbf{told}!                    & Imagine you \textbf{tell} someone \textit{sago} is good for them.           \\
                   & Friseur  & Imagine a \textbf{hairdresser} who can tame even the most \textit{frizzy} hair!               & Imagine your \textbf{hairdresser} inside a \textit{freezer}.                \\\hline
\multirow{2}{*}{C} & Nehmen   & Imagine \textit{Newman} taking the initiative to \textbf{take} action!                        & Imagine you \textbf{take} a \textit{name} in your address book.             \\
                   & Brauchen & Imagine \textbf{needing} to fix a \textit{broken} heart.                                      & Imagine \textit{brokers} \textbf{need} much experience. \\\hline                   
\end{tabularx}
\label{tab:verbal_cue}
\end{table*}

\subsubsection{Auto-II vs.\ Auto-III: Does a visual cue help learning?}
We hypothesize better performance by Auto-III, which uses additional visual cues, than Auto-II. Therefore, we define our null hypothesis and alternate hypothesis as follows:
\begin{itemize}
  \item $H_{0}$: $\mu_{Auto-III} \leq \mu_{Auto-II}$
  \item $H_{a}$: $\mu_{Auto-III} > \mu_{Auto-II}$
\end{itemize}

A right-tailed test shows that there is no significant difference between the two conditions, $t(32) = 0.39, p = 0.35$; we cannot reject $H_{0}$. In Fig.~\ref{fig:combined_score}, we show three words for the cases where participants perform better in the Auto-III condition than in Auto-II (case B) and two for when it does not (case C), respectively. Case B shows that Auto-III, which has additional visual cues than Auto-II, performs similarly as Manual-II. Considering the previous comparison that Auto-II has a lower score than Manual-II, we see that Auto-III does somewhat outperform Auto-II. Therefore, we can conclude that visual cues help participant build the imagery link to some degree. 

\begin{figure}
\captionsetup[subfigure]{}
\begin{tabular}{c c c c}
Case B 
& 
\begin{subfigure}{0.3\textwidth}
\centering
\includegraphics[width=.65\linewidth]{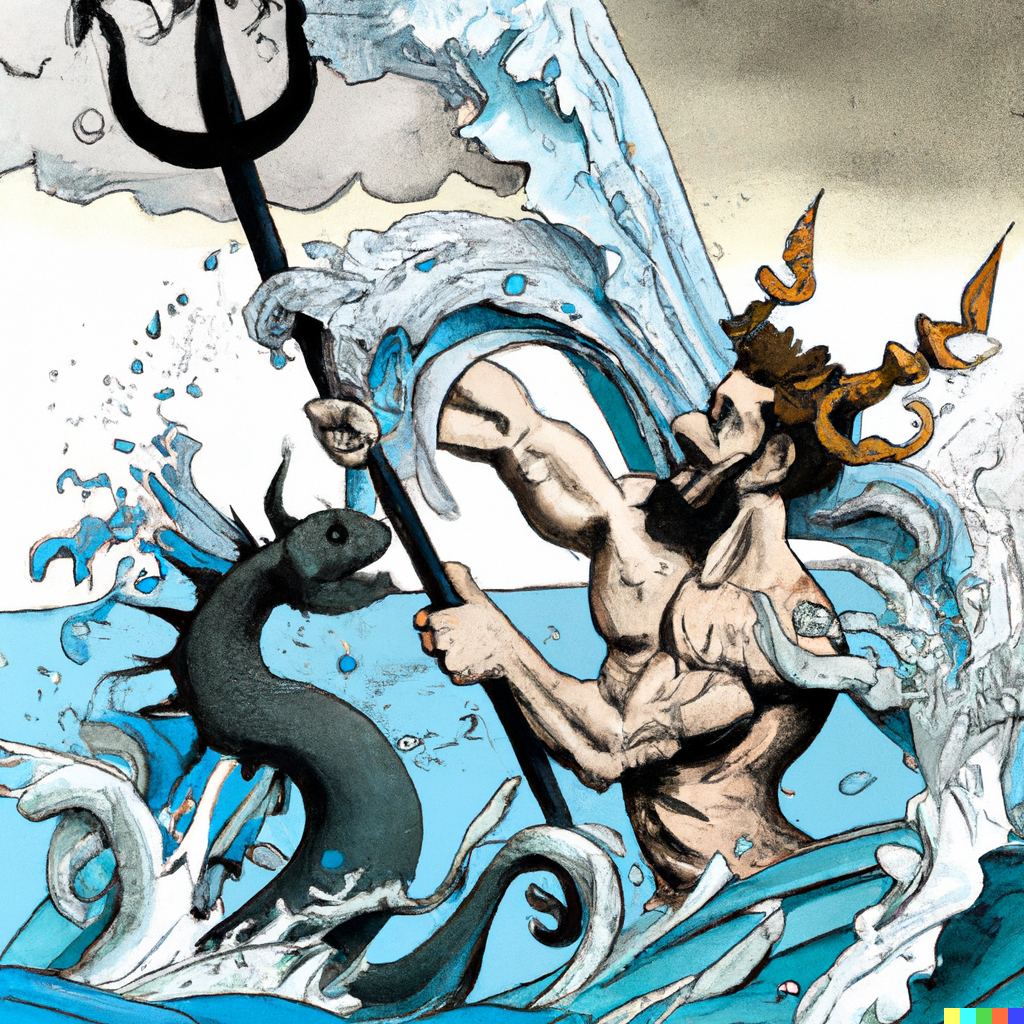}
\caption{Streiten}
\end{subfigure}%
&
\begin{subfigure}{0.3\textwidth}
\centering
\includegraphics[width=.65\linewidth]{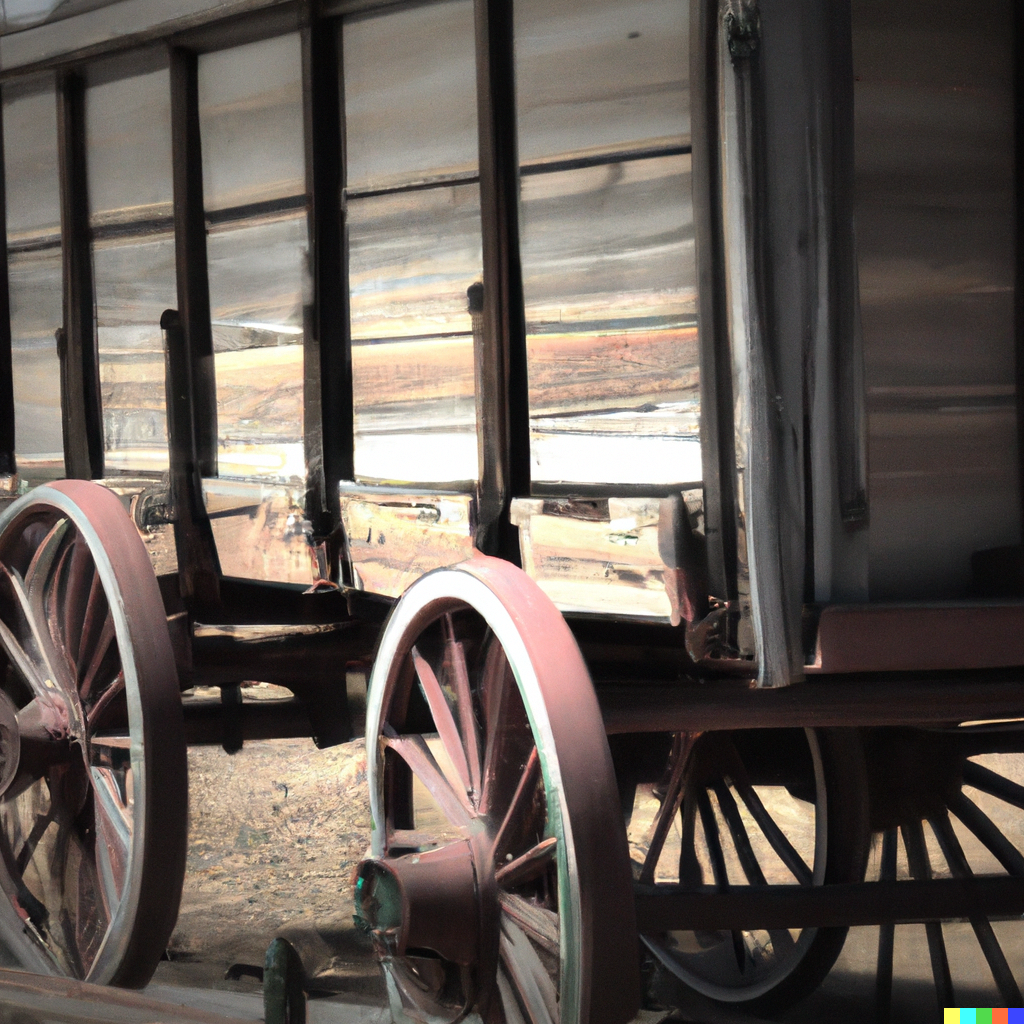}
\caption{Sagen}
\end{subfigure}%
& 
\begin{subfigure}{0.3\textwidth}
\centering
\includegraphics[width=.65\linewidth]{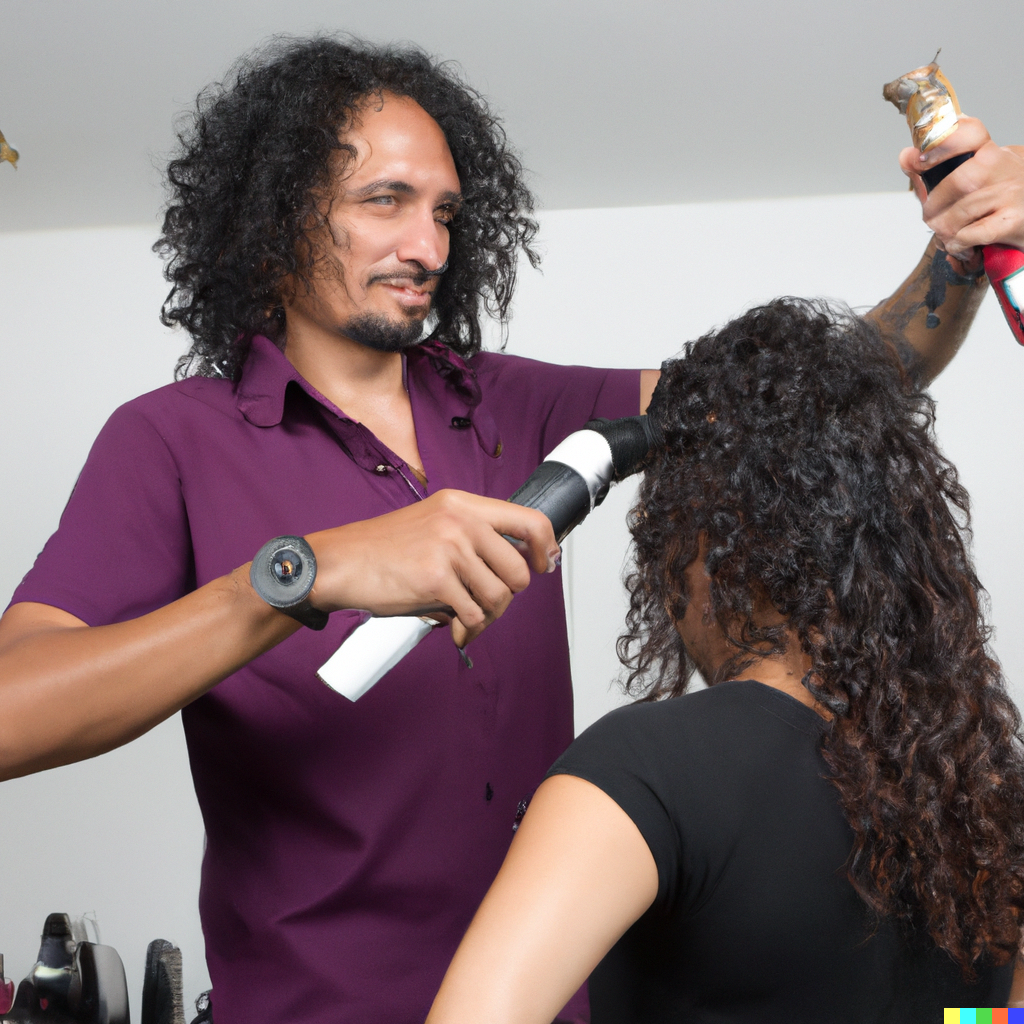}
\caption{Friseur}
\end{subfigure}
\end{tabular}
\begin{tabular}{c c c}
Case C
&
\begin{subfigure}{0.3\textwidth}
\centering
\includegraphics[width=.65\linewidth]{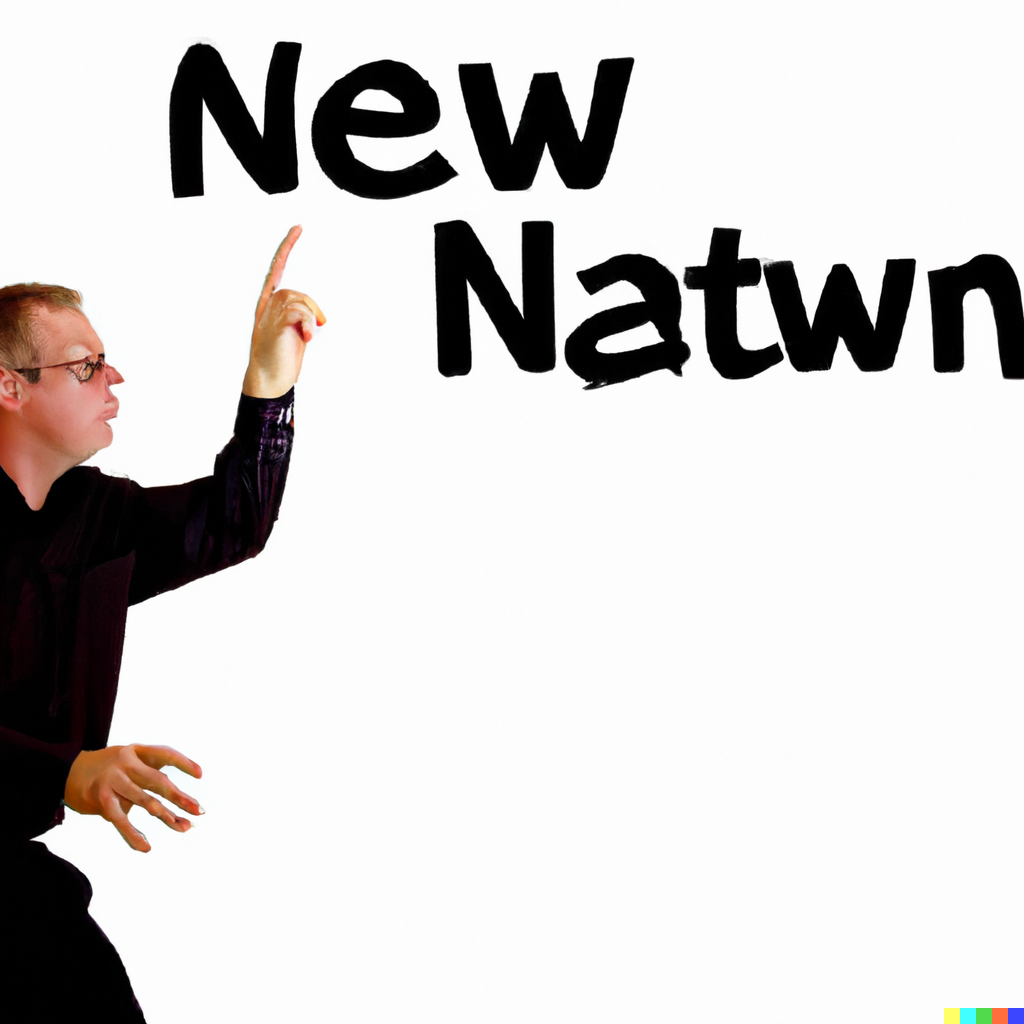}
\caption{Nehmen}
\end{subfigure}%
&
\begin{subfigure}{0.3\textwidth}
\centering
\includegraphics[width=.65\linewidth]{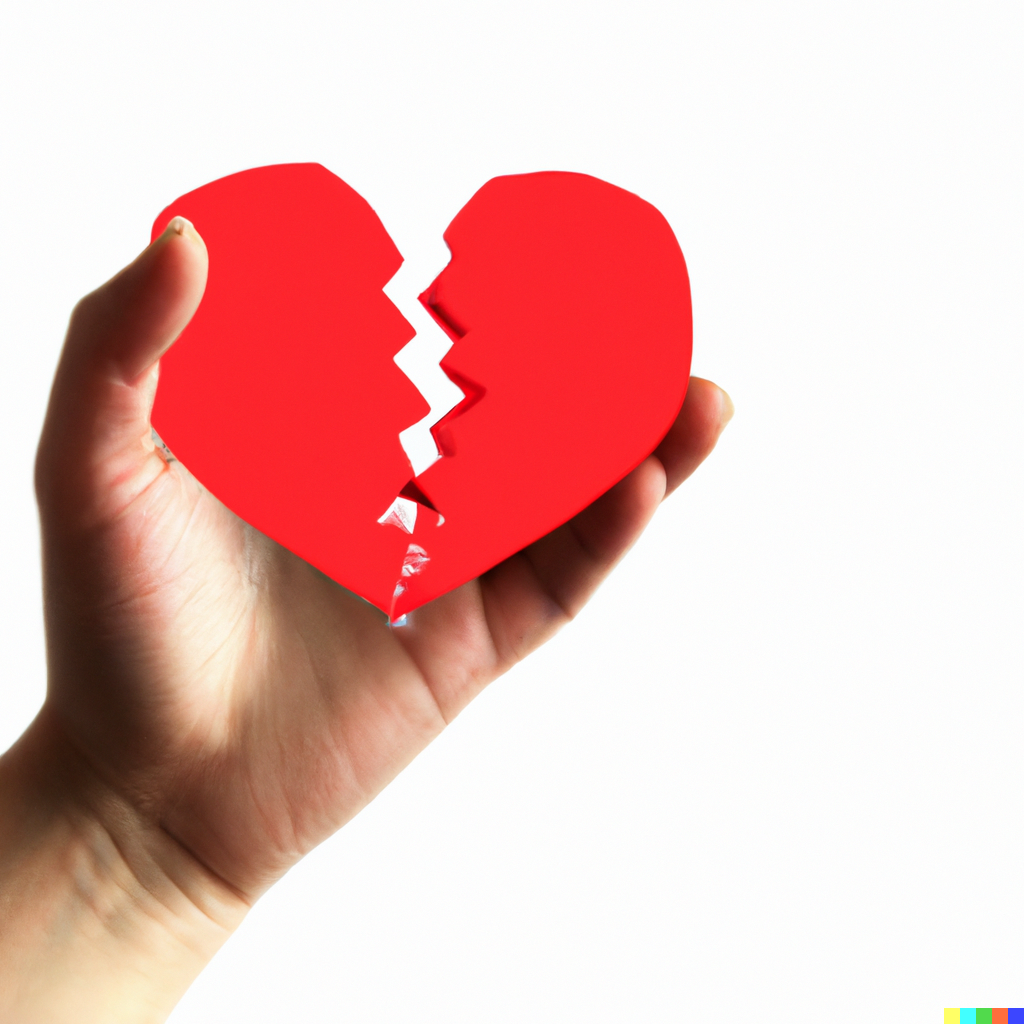}
\caption{Brauchen}
\end{subfigure}%
\end{tabular}
\caption{Examples of visual cues generated by our pipeline in cases where they are helpful to participants and cases where they are not.}
\label{fig:visual_cue}
\vspace{-.5cm}
\end{figure}

For a more qualitative analysis, Fig.~\ref{fig:visual_cue} shows visual cues generated by our pipeline. Fig.~\ref{fig:visual_cue} (a-c) shows that visual cues may be helpful in cases where keywords that lack imageability and are not frequently used (Triton, frizzy) or in cases where auto-generated verbal cues are hard to imagine (a wagon with stories). However, as shown in case C, visual cues for abstract words (to take, to need) do not help much. Fig.~\ref{fig:visual_cue} (d-e) shows that in these cases the generated image is not descriptive enough to facilitate the imagery link. Interestingly, the Likert scale score was higher for Auto-III than Auto-II in every word except one. This result implies that participants think it is helpful to have additional visual cues. However, we cannot create effective visual cues for every word. Generating descriptive visual cues, especially for abstract words, remains a challenging task. 

\section{Conclusions and Future Work}

In this paper, we explored the opportunity of using large language models to generate verbal and visual cues for keyword mnemonics. A preliminary human experiment suggested that despite showing some promise, this approach has limitations and cannot reach the performance of manually generated cues yet. 

There are many avenues for future work. First, we need a larger-scale experiment in a real lab study, which provides us a controlled environment to test both short-term and long-term retention. Since we only tested short-term retention, it is possible that no approach can significantly outperform others. We also need more psycholinguistics perspectives on constraining time spent on learning and testing. By conducting the research in a more controlled environment, we can use additional information (e.g., demographics, language level) to help us conduct a deeper analysis of the results. We do clarify that using Amazon's Mechanical Turk to conduct experiments is standard in prior work, which is part of the reason why we chose this experimental setting. To track long-term retention, we likely have to resort to knowledge tracing models that handle either memory decay \cite{akt} or open-ended responses \cite{okt}. 
Second, we can extend our pipeline by generating the keyword automatically as well instead of using TransPhoner-generated keywords, which may make our approach even more scalable. One important aspect that must be studied is how to evaluate the imageability of the keywords and verbal cue that contains both keywords and vocabulary, which remains challenging. 
Third, we can generate personalized content for each participant. We may provide additional information in the text generator about the topic they are interested in that we could use to generate a verbal cue. Moreover, we can generate a story that takes all words into account. It is also possible to generate verbal cues in L2 as well, which may help learners by providing even more context.
\begin{CJK*}{UTF8}{gbsn}
Fourth, instead of the pronunciation of the word, we can use other features in language to generate verbal cues. For example, when learning Mandarin, memorizing Chinese characters is as important as learning how to pronounce the word.
The Chinese character 休\phantom{} means \textit{rest}, which is xiū in Mandarin. The character is called a compound ideograph, a combination of a person (人) and a tree (木), which represents a person resting against a tree. Combined with a keyword, \textit{shoe}, for example, we could accomplish two goals with one verbal cue, ``A person is \textit{resting} by a tree, tying up their \textit{shoe}.'' This way, we can make visual cues more descriptive for abstract words. 
\end{CJK*}

\section{Acknowledgements}
The authors thank the NSF (under grants 1917713, 2118706, 2202506, 2215193) for partially supporting this work. 

\bibliographystyle{splncs04}
\bibliography{main}

\begin{thebibliography}{10}
\providecommand{\url}[1]{\texttt{#1}}
\providecommand{\urlprefix}{URL }
\providecommand{\doi}[1]{https://doi.org/#1}

\bibitem{duolingo}
Ahn, L.v.: Duolingo, \url{https://www.duolingo.com}

\bibitem{amt}
Amazon: Amazon mechanical turk, \url{https://www.mturk.com}

\bibitem{atkinson1975application}
Atkinson, R.C., Raugh, M.R.: An application of the mnemonic keyword method to
  the acquisition of a russian vocabulary. Journal of experimental psychology:
  Human learning and memory  \textbf{1}(2), ~126 (1975)

\bibitem{bojanowski2017enriching}
Bojanowski, P., Grave, E., Joulin, A., Mikolov, T.: Enriching word vectors with
  subword information. Transactions of the association for computational
  linguistics  \textbf{5},  135--146 (2017)

\bibitem{brahler2008learning}
Brahler, C.J., Walker, D.: Learning scientific and medical terminology with a
  mnemonic strategy using an illogical association technique. Advances in
  physiology education  \textbf{32}(3),  219--224 (2008)

\bibitem{brown2020language}
Brown, T., Mann, B., Ryder, N., Subbiah, M., Kaplan, J.D., Dhariwal, P.,
  Neelakantan, A., Shyam, P., Sastry, G., Askell, A., et~al.: Language models
  are few-shot learners. Advances in neural information processing systems
  \textbf{33},  1877--1901 (2020)

\bibitem{carrier1992influence}
Carrier, M., Pashler, H.: The influence of retrieval on retention. Memory \&
  cognition  \textbf{20},  633--642 (1992)

\bibitem{ebbinghaus2013memory}
Ebbinghaus, H.: Memory: A contribution to experimental psychology. Annals of
  neurosciences  \textbf{20}(4), ~155 (2013)

\bibitem{ellis1993psycholinguistic}
Ellis, N.C., Beaton, A.: Psycholinguistic determinants of foreign language
  vocabulary learning. Language learning  \textbf{43}(4),  559--617 (1993)

\bibitem{anki}
Elmes, D.: Anki, \url{http://ankisrs.net}

\bibitem{akt}
Ghosh, A., Heffernan, N., Lan, A.S.: Context-aware attentive knowledge tracing.
  In: Proc. ACM SIGKDD. pp. 2330--2339 (2020)

\bibitem{larsen2009repeated}
Larsen, D.P., Butler, A.C., Roediger~III, H.L.: Repeated testing improves
  long-term retention relative to repeated study: a randomised controlled
  trial. Medical education  \textbf{43}(12),  1174--1181 (2009)

\bibitem{leitner1974so}
Leitner, S.: So lernt man lernen. Herder (1974),
  \url{https://books.google.com/books?id=opWFRAAACAAJ}

\bibitem{okt}
Liu, N., Wang, Z., Baraniuk, R., Lan, A.: Open-ended knowledge tracing for
  computer science education. In: Conference on Empirical Methods in Natural
  Language Processing. pp. 3849--3862 (2022)

\bibitem{dalle}
OpenAI: Dall-e 2, \url{https://openai.com/dall-e-2}

\bibitem{ouyang2022training}
Ouyang, L., Wu, J., Jiang, X., Almeida, D., Wainwright, C., Mishkin, P., Zhang,
  C., Agarwal, S., Slama, K., Ray, A., et~al.: Training language models to
  follow instructions with human feedback. Advances in Neural Information
  Processing Systems  \textbf{35},  27730--27744 (2022)

\bibitem{prabhumoye2020exploring}
Prabhumoye, S., Black, A.W., Salakhutdinov, R.: Exploring controllable text
  generation techniques. arXiv preprint arXiv:2005.01822  (2020)

\bibitem{reddy2016unbounded}
Reddy, S., Labutov, I., Banerjee, S., Joachims, T.: Unbounded human learning:
  Optimal scheduling for spaced repetition. In: Proceedings of the 22nd ACM
  SIGKDD international conference on knowledge discovery and data mining. pp.
  1815--1824 (2016)

\bibitem{savva2014transphoner}
Savva, M., Chang, A.X., Manning, C.D., Hanrahan, P.: Transphoner: Automated
  mnemonic keyword generation. In: Proceedings of the SIGCHI Conference on
  Human Factors in Computing Systems. pp. 3725--3734 (2014)

\bibitem{siriganjanavong2013mnemonic}
Siriganjanavong, V.: The mnemonic keyword method: Effects on the vocabulary
  acquisition and retention. English Language Teaching  \textbf{6}(10),  1--10
  (2013)

\bibitem{quizlet}
Sutherland, A.: Quizlet, \url{http://quizlet.com}

\bibitem{ye2022stochastic}
Ye, J., Su, J., Cao, Y.: A stochastic shortest path algorithm for optimizing
  spaced repetition scheduling. In: Proceedings of the 28th ACM SIGKDD
  Conference on Knowledge Discovery and Data Mining. pp. 4381--4390 (2022)

\bibitem{zylich2021linguistic}
Zylich, B., Lan, A.: Linguistic skill modeling for second language acquisition.
  In: LAK21: 11th International Learning Analytics and Knowledge Conference.
  pp. 141--150 (2021)

\end{thebibliography}
\end{document}